\documentclass{article}

\usepackage[nonatbib,preprint]{neurips_2020}

\usepackage[utf8]{inputenc} % allow utf-8 input
\usepackage[T1]{fontenc}    % use 8-bit T1 fonts
\usepackage{hyperref}       % hyperlinks
\usepackage{url}            % simple URL typesetting
\usepackage{booktabs}       % professional-quality tables
\usepackage{amsfonts}       % blackboard math symbols
\usepackage{nicefrac}       % compact symbols for 1/2, etc.
\usepackage{microtype}      % microtypography
\usepackage{graphicx}
\usepackage{bm}
\title{Condensing Two-stage Detection with Automatic Object Key Part Discovery}

\author{%
 Zhe Chen ~ Jing Zhang ~ Dacheng Tao\\
  UBTECH Sydney AI Centre\\
 School of Computer Science, Faculty of Engineering, The University of Sydney,\\
  Australia \\
  \texttt{ \{zhe.chen1, jing.zhang1, dacheng.tao\}@sydney.edu.au} \\
}

\begin{document}

\maketitle

\begin{abstract}
Modern two-stage object detectors generally require excessively large models for their detection heads to achieve high accuracy. 
To address this problem, we propose that the model parameters of two-stage detection heads can be condensed and reduced by concentrating on object key parts. To this end, we first introduce an automatic object key part discovery task to make neural networks discover representative sub-parts in each foreground object. With these discovered key parts, we then decompose the object appearance modeling into a key part modeling process and a global modeling process for detection. Key part modeling encodes fine and detailed features from the discovered key parts, and global modeling encodes rough and holistic object characteristics. In practice, such decomposition allows us to significantly abridge model parameters without sacrificing much detection accuracy. Experiments on popular datasets illustrate that our proposed technique consistently maintains original performance while waiving around 50\% of the model parameters of common two-stage detection heads, with the performance only deteriorating by 1.5\% when waiving around 96\% of the original model parameters. Codes are released on: \url{https://github.com/zhechen/Condensing2stageDetection}.
%We observe that these detectors exhaustively model all the appearance details to detect objects, while we humans can accurately detect objects by only focusing on some key parts. 
%and then seek to alter and reduce model capacities for existing detection heads based on the discovered object key parts, revealing a great potential of the proposed method to save model consumption of two-stage detectors
\end{abstract}

\section{Introduction}
Object detection is a key task in computer vision. By identifying and locating objects belonging to certain classes in an image, object detectors are critical in various applications like autonomous driving \cite{geiger2013vision}, traffic surveillance \cite{dollar2009pedestrian,dollar2012pedestrian}, tracking \cite{voigtlaender2019mots} and so on. 
Currently, two-stage deep convolutional neural network (DCNN) based detectors can be used for high-quality detection. However, accurate models \cite{felzenszwalb2010cascade,chen2019hybrid} commonly require a large detection head with excessive parameters to achieve robust classification and regression based on extracted features, making existing state-of-the-art two-stage detectors less practical for platforms with limited memory. 

Several attempts have been made to reduce model parameters of two-stage detectors. For example, single-stage detectors \cite{duan2019centernet,redmon2016you,liu2016ssd} have been developed to merge the detection head into the backbone network. However, it can be challenging to achieve state-of-the-art performance with smaller models of these detectors. The detectors winning recent major competitions like MS COCO \cite{lin2014microsoft} are still two-stage or even multi-stage\cite{liu2018path,chen2019hybrid}. Efficient light-weight detectors \cite{hong2016pvanet,wang2018pelee,qin2019thundernet} have also been devised to simplify detection, but, similar to single-stage detectors, these detectors tend not to achieve comparable performance to cutting-edge two-stage detectors. Moreover, recent progress in network compression \cite{han2015deep,wu2016quantized,li2016pruning} has opened up the possibility of compressing detection models. However, most of the existing approaches that compress two-stage detectors sacrifice the performance significantly. Some effective approaches \cite{singh2019multi,wang2019distilling} only work for single-stage detectors like SSD\cite{liu2016ssd}. 
Alternatively, we propose a novel technique that performs detection based on object key parts, thereby effectively reducing the model parameters of the two-stage detection heads required to achieve high detection accuracy. We term this technique as "\emph{condensing}" two-stage detection. 

\begin{figure}[t]
\includegraphics[width=\textwidth,height=3.2cm]{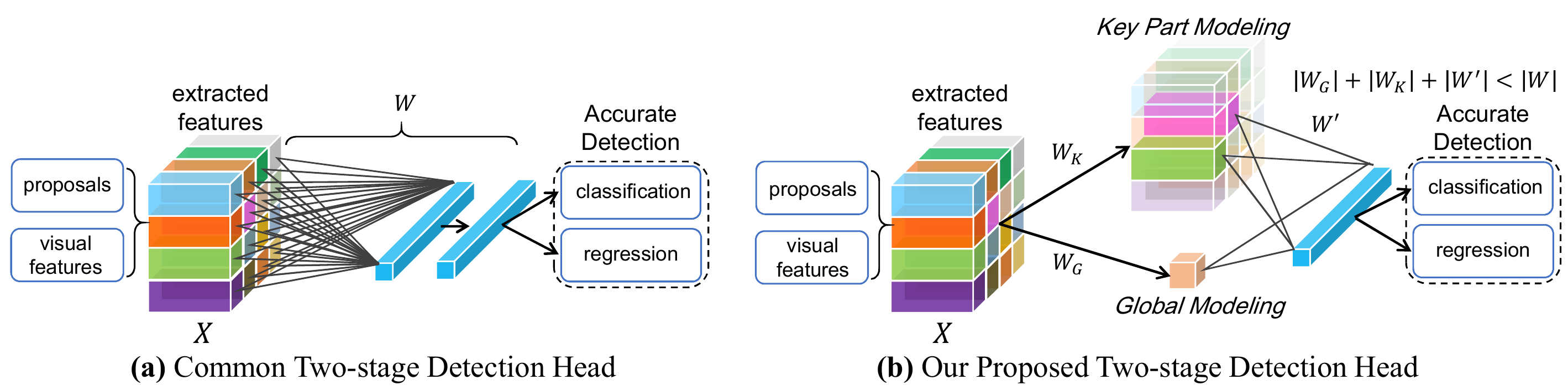}
  \caption{Common two-stage detection head \textbf{(a)} compared to our proposed detection head \textbf{(b)} that condenses and reduces the model parameters required for accurate detection based on key parts. $W$, $W_{KR}$, $W_{CR}$, $W'$ are all network parameters. Best view in color.}%
  \label{fig:intro}
\vspace{-0.5cm}
\end{figure}

%In general,
%Accordingly, we condense the two-stage detection heads by focusing on the modeling of a few key parts and holistic appearances for detection. 
%To effectively condense two-stage detection with object key parts, 
Existing two-stage detection heads generally model every detailed feature extracted for detection. We humans, on the other hand, can detect an object without perceiving all of its details. Early biological studies \cite{bar1996spatial,selfridge1955pattern,tanaka1993parts} revealed that we could recognize an item based on its parts and whole appearance and can robustly perform detection if only some key parts are visible. Motivated by these findings, we propose that accurate detection can be approached in computer vision using object key parts and holistic appearance rather than modeling every feature, thus avoiding the need for excessive parameters. 
However, to our best knowledge, no existing two-stage DCNN detectors can discover object key parts and condense detection heads using these key parts. 
To tackle this problem, we first introduce a novel learning task, automatic object key part discovery, to help networks find appropriate key parts autonomously.
Using discovered key parts, we then recast two-stage detection heads to reduce the model parameters for accurate detection. Figure \ref{fig:intro} compares the common two-stage detection heads and our proposed approach.

% to help DCNNs able to perform automatic object key part discovery and to achieve robust detection only based on identified key parts. We name the proposed head as Object Key Part Discovery Network (OKPD-Net).
%By adequately constraining confidence values and training OKPD-Net, we can obtain a set of discovered key parts that can better represent objects.
%In the OKPD-Net, we introduce a truncated maximum regularization to better squash the produced confidence values, and we apply a discriminative loss and a uniqueness loss to achieve more adequate training. 
More specifically, we first formulate automatic object key part discovery as the task of making neural networks automatically and autonomously discover representative sub-parts of an object. This automatic discovery avoids the need for labor-costly manual annotation to define key parts. The discovered key parts of each object can then help condense two-stage detection heads by waiving the modeling of non-key parts. In particular, we employ a small neural network, called the Object Key Part Discovery Network (OKPD-Net), to complete this new task by producing confidence maps for different key parts. 
With these discovered key parts, we then condense two-stage detection heads by decomposing the object appearance modeling into a key part modeling process and a global modeling process. Key part modeling collects and models visual features from the discovered key parts, providing essential fine details to access accurate detection. 
Global modeling is similar to how humans glimpse at objects. It extracts and models rough and holistic descriptions about objects, estimating their general states to help achieve robust detection.

To sum up, our main contributions are as follows:
\begin{itemize}
    \item We introduced a novel technique that uses object key parts to effectively reduce model parameters required for two-stage detection heads without greatly deteriorating performance. 
    \item  A novel learning task, automatic object key part discovery, is formulated to help networks discover representative sub-parts of objects autonomously. To condense model parameters using key parts, we innovatively decompose the object appearance modeling into a key part modeling process and a global modeling process.
    \item Our technique can effectively condense and reduce the model parameters of various two-stage detection heads without losing much detection accuracy. In particular, our approach can waive approximately \emph{50\%} of original model parameters while retaining the original performance on popular object detection benchmarks like PASCAL VOC \cite{everingham2010pascal} and MS COCO\cite{lin2014microsoft}. Furthermore, when waving \emph{96\%} of the original model parameters, our approach is still highly accurate, suffering only a \emph{1.5\%} drop in performance \textit{w.r.t.} the baseline model, demonstrating its effectiveness in saving model parameters for two-stage detection. 
\end{itemize}

%In the rest of this paper, we will subsequently describe our processing framework, the automatic key part discovery task, the design and training of OKPD-Net, and how we decompose the modeling of object appearance to condense two-stage detection heads in Section \ref{sec:3}. Quantitative and qualitative results are illustrated in Section \ref{sec:4}. 

\vspace{-0.2cm}
\section{Related Work}
\vspace{-0.2cm}
\textbf{Single-stage Object Detection}
Two-stage detection first extracts proposals and then performs recognition and localization for detection \cite{girshick2014rich,ren2015faster}. Conversely, single-stage detectors  \cite{redmon2016you,redmon2017yolo9000,liu2016ssd,duan2019centernet,law2018cornernet} directly recognize and localize objects without proposals. This enables these detectors to merge their detection heads within the backbone network to save computational resources. For example, Yolo \cite{redmon2016you} and SSD \cite{liu2016ssd}, two of the most popular single-stage detectors, have less model parameters and could reach real-time processing speed. Howeveer, single-stage detectors are generally difficult to achieve comparable performance with cutting-edge two-stage detectors using small models. They still need a large number of parameters, like RetinaNet \cite{lin2017focal}, to achieve high accuracy. On the other hand, recent winning detection technologies \cite{liu2018path,chen2019hybrid} in MS COCO \cite{lin2014microsoft} are still two-stage or multi-stage. 

\textbf{Light-weight Model and Network Compression}
Another direction for detection with fewer model parameters is to introduce light-weight architectures\cite{hong2016pvanet,wang2018pelee,qin2019thundernet}. For example, both PVA Net\cite{hong2016pvanet} and VoVNet \cite{lee2019energy,lee2019centermask} use smaller networks to reduce model parameters used in detection, but they would cause obvious performance deterioration after applying lighter-weight models. Besides, they still rely on full appearance details for detection, which also needs excessive parameters.
%\textbf{Network Compression}
Recent progress in network pruning\cite{han2015deep,wu2016quantized,li2016pruning} or knowledge distillation\cite{bucilua2006model,hinton2015distilling} has effectively compressed classification networks, encouraging attempts to also compress detection models. However, very few successful techniques can effectively reduce model parameters of object detectors without significant performance loss. To our best knowledge, most of effective studies like \cite{singh2019multi,chu2019mixed} only work on single-stage detectors. For the two-stage detectors, studies on pruning \cite{He_2017_ICCV} or knowledge distillation \cite{chen2017learning,wang2019distilling} achieve some progress, but they reduce the parameters at the obvious cost of accuracy. 

\textbf{Part-based and Attention-based Detection}
Both part-based models and attention-based models also put more focus on key areas for detection. Part-based models like \cite{felzenszwalb2010object,zhang2014part,mordan2019end} are effective for detection and recognition, but they are commonly dedicated to improve performance by introducing extra complexity to model parts. The used parts usually serve as additional information sources rather than better representing objects and discriminating them from background.
Besides, attention-based models \cite{zhang2018progressive,wang2018non} can learn to find representative areas to facilitate detection, but they still tend to introduce large models to encode every details. 

\vspace{-0.2cm}
\section{Condensing Two-stage Detection with Automatic Object Key Part Discovery}
\label{sec:3}
\vspace{-0.2cm}
%In this study, we propose that model capacities of two-stage detection heads can be effectively condensed and reduced by introducing automatic object key parts. This section will describe in details about the proposed technique. 

\begin{figure}[t]
\includegraphics[width=0.85\textwidth,height=4.3cm]{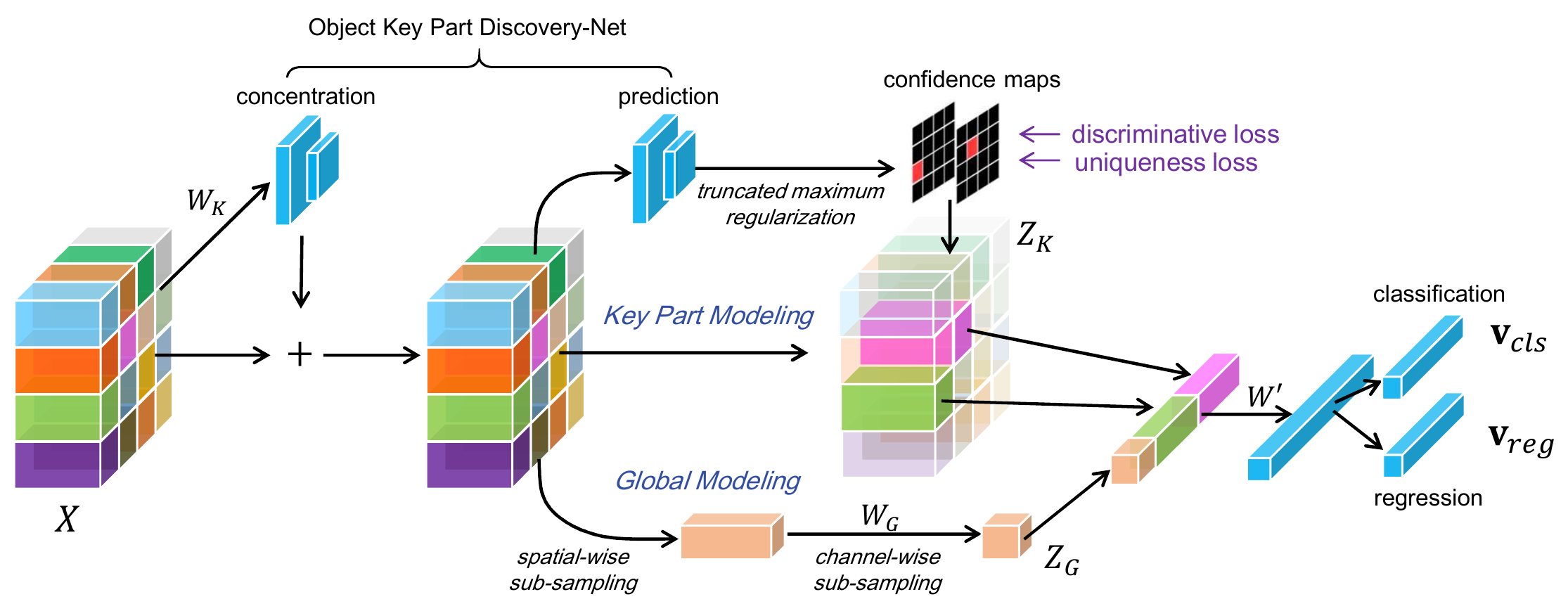}
  \caption{Overall processing framework of our proposed detection head. We first employ an object key part discovery network to refine input features and learn to predict key parts automatically. With the predicted key parts, we perform key part modeling and global modeling for detection, which condenses and reduces model parameters required for accurate detection.}%
  \label{fig:framework}
\vspace{-0.5cm}
\end{figure}

\vspace{-0.1cm}
\subsection{Processing Framework}
\label{sec:3-1}
\vspace{-0.1cm}
In general, current two-stage object detectors first generate region proposals \cite{uijlings2013selective,girshick2014rich}, each of which may potentially contain an object. Then, convolutional features within each proposal are extracted and fed into a detection head that extracts semantics and performs detection (classification and regression) to predict object classes and spatial offsets, respectively. 

Formally, we denote $X$ as the collection of features extracted from a proposal: $X = \{\mathbf{x}_1, \ldots, \mathbf{x}_i, \ldots, \mathbf{x}_N\}$, where $\mathbf{x}_i$ is $i$-th feature vector and $N$ represents the total number of elements in $X$. In many two-stage detectors, $X$ is obtained by sampling visual features within the proposal window into a tensor whose spatial size is 7 by 7, \textit{i.e.} $N=49$ in this case. Suppose $\mathbf{v}_{cls}$ and $\mathbf{v}_{reg}$ represent classification and regression results, respectively. Then, the two-stage detection heads can be described by:
\begin{equation}
\mathbf{v}_{cls}, \mathbf{v}_{reg} = f(X; W),
\label{eq:orig}
\end{equation}
where $f(\cdot)$ represents the network that maps $X$ to $\mathbf{v}_{cls}$ and $\mathbf{v}_{reg}$ using a parameter set $W$. Fully connected layers are usually applied to implement $f$.
%In this study, we propose that using all the 7 by 7 features from $X$ could introduce an excessive amount of weight parameters, thus result in larger model capacities. 
By contrast, we introduce object key parts to help condense and reduce the parameters required for modeling $X$. 

Figure \ref{fig:framework} shows an overview of the proposed detection head. First, we consider an object part as a sub-region which provides a feature vector in $X$.
We introduce automatic object key part discovery to let the employed network learn to discover an adequate key part set, denoted as $K$, for detection. Each element in $K$ describes a location on the tensor of $X$ where the corresponding feature can better represent objects. Then, we decompose the modeling of $X$ into a key part modeling process and a global modeling process. Based on $K$, we formulate the key part modeling process as to model features from key parts. Meanwhile, in the global modeling process, we sub-sample $X$ both spatial-wise and channel-wise to provide rough and holistic cues about input objects. We use $Z_{K}$ and $Z_{G}$ to describe the results of key part modeling and global modeling, respectively.
In summary, we condense two-stage detection heads based on:
\begin{equation}
\mathbf{v}_{cls}, \mathbf{v}_{reg} = f(Z_{K}, Z_{G}; W'), 
\label{eq:overall}
\end{equation}
where $W'$ is the new network parameter set for mapping $Z_{K}$ and $Z_{G}$ to $\mathbf{v}_{cls}$ and $\mathbf{v}_{reg}$. We mainly use one fully connected layer for $f$, saving around 7\% parameters compared to the two-layer network.
The $Z_{K}$ and $Z_{G}$ are respectively obtained by:
\begin{equation}
Z_{K} = f_{K}(X;W_{K}), ~~~ Z_{G} = f_{G}(X;W_{G}),
\label{eq:xd}
\end{equation}
where $f_{K}$ is the function that implements key part modeling, $f_{G}$ is the function that implements global modeling, and $W_{K}$ and $W_{G}$ are their corresponding network parameters. 

In this paper, we use $|W|$ to denote the number of elements in $W$. Therefore, this study seeks to use a smaller overall parameter set, \textit{i.e.} $|W'| + |W_{K}| + |W_{G}| < |W|$, to achieve high-quality detection. 

%In the following sections, we will subsequently discuss more comprehensively about how we implement $f_{KR}$ and $h$ to decompose $X$ and reduce model capacities without losing too much detection performance. 

% neural networks automatically discover key parts of an object and then reduce the scale of $X$ and further reduce the model parameters required for accurate estimation of $\mathbf{v}_{cls}$ and $\mathbf{v}_{reg}$. 

\vspace{-0.1cm}
\subsection{Automatic Object Key Part Discovery}
\label{sec:3-2}
\vspace{-0.1cm}
\subsubsection{Task Description}
\vspace{-0.1cm}
When training a network to discover key parts adequately, it is difficult to define what shoud be regarded as key parts of objects belonging to a wide range of classes. Annotating object key parts is itself a significantly labor-costly task. To avoid such difficulties, 
we formulate an automatic object key part discovery task that aims to make neural networks autonomously learn to discover the representative sub-parts of an object. To fulfill this task, we employ a small neural network, the object key part discovery network (OKPD-Net), to estimate key part locations.
%Then, we introduce proper training objectives to help OKPD-Net provide adequate estimation of the locations of object key parts.  

%It is analogous to our human learning process, in which we can learn to identify the most representative parts of an object only by ourselves. 
%In addition, the discovered key sub-parts of an object can later facilitate the recognition and localization process without introducing too much extra model capacities.

\vspace{-0.1cm}
\subsubsection{Network Design}
\vspace{-0.1cm}
In OKPD-Net, we perform two processing steps, a concentration step and a prediction step. The concentration step consists of several small kernel-based convolutions to help the network focus on more beneficial cues for detection. Then, in the prediction step, we make OKPD-Net produce different confidence maps to represent discovered key parts. 
%By adequately constraining the produced confidence values, we can consider the sub-part with maximum confidence value on each map as the object key part. 
%Figure \ref{} shows the architecture of the employed OKPD-Net. 

In the concentration step, we apply two convolution blocks. Each convolutional block consists of a 3 by 3 dilated group convolution with small channel numbers, \textit{i.e.,} eight times smaller than the input channel. We then apply a 1 by 1 convolution to restore the reduced channels. The output of each block is added to its input, forming a residual structure. 

After the concentration step, we apply a 1 by 1 convolution to predict different confidence maps for corresponding key parts. To better represent and interpret confidence, the predicted values should be within an adequate range, but the common squashing functions like sigmoid or softmax are sub-optimal for this task: the sigmoid function ignores the competition between the maximum value and other values, and the softmax function usually produces small initial values and requires supervision to highlight the desired output. However, we prefer the highlighted maximum confidences to reveal the key parts without relying on annotations. To this end, we devise a novel function, truncated maximum regularization, to better constrain and squash the confidence values for the task.

\textbf{Truncated Maximum Regularization} For each confidence map, the truncated maximum regularization function normalizes confidence values within the range [0,1) by dividing each value by the maximum value on that map when the maximum value surpasses 1. Suppose $c$ refers to each value on a confidence map and $c_{m}$ is the maximum value on that map. The truncated maximum regularization squashes $c$ according to:
\begin{equation}
c = max\{0, (c+\alpha) / \big(max\{0, (c_{m}+\alpha) - 1\} +1 + \epsilon\big) \},
\label{eq:sq}
\end{equation}
where $\alpha$ is an offset value and $\epsilon$ is a small value. The offset value is introduced to set initial values of $c$, since $c$ is usually around 0 at the start of training. We empirically set $\alpha$ to 0.5. The $\epsilon$ then prevents the squashed maximum confidence becoming the constant value 1, which can better help back-propagate information. $\epsilon$ is set to 0.1 in this study. As a result, the proposed function in Eq. \ref{eq:sq} is initially a linear function, which is easier for training, and it can also make $c_{m}$ compete with other values when $c_{m}$ becomes large. 

\vspace{-0.1cm}
\subsubsection{Training Objective}
\vspace{-0.1cm}
The training objective is critical to tell OKPD-Net which parts are favored and what parts are not. We introduce the training objective, denoted $\mathcal{L}$, to train OKPD-Net. In this study, we apply a discriminative loss $\mathcal{L}_d$ to help OKPD-Net explore more representative key parts and also apply a uniqueness loss $\mathcal{L}_u$ to prevent predicting overlapping key parts. $\mathcal{L}$ can thus be written as:
\begin{equation}
\mathcal{L} = \mathcal{L}_d + \mathcal{L}_u.
\end{equation}

%\vspace{-0.1cm}
\textbf{Discriminative Loss} 
%\vspace{-0.1cm}
The difficulty in training OKPD-Net is that we do not have a clear definition of what defines a key part of an object. Instead of introducing full supervision which is strict and labor costly, we tend to introduce a softer training criteria which allows to predict any local part of an object as a key part so long as it better represents foreground objects. Accordingly, by making full use of the available bounding box annotations, we introduce a discriminative loss $\mathcal{L}_d$ to enhance the key part confidences for foreground objects and suppress the confidences for background areas. 
% to ensure that the key parts predicted by OKPD-Net can better represent foreground objects. 

Formally, we denote $c^{i,k}_{m}$ as the maximum confidence value of the $k$-th key part from the $i$-th input example, and denote $\hat{y}^{i}$ as the label of the $i$-th input example, indicating it properly contains a foreground object: $\hat{y}^{i} = 1$ if the $i$-th input example contains and $\hat{y}^{i} = 0$ if not. Thus, we have $\mathcal{L}_d$ as:
\begin{equation}
\mathcal{L}_d = \sum_{i,k} l[c^{i,k}_{m}, \hat{y}^{i}],
\label{eq:ld}
\end{equation}
where $l$ is the distance between $c^{i,k}_{m}$ and $\hat{y}^{i}$. Considering $c$ is computed based on Eq. \ref{eq:sq}, we find that the smooth L1 loss \cite{ren2015faster} is more appropriate for defining $l$. As a result, Eq. \ref{eq:ld} describes that the maximum value on each key part confidence map should be 1 for positive examples and \textit{vice versa}. 

%\vspace{-0.1cm}
\textbf{Uniqueness Loss} 
%\vspace{-0.1cm}
The discriminative loss alone can already train OKPD-Net, but the predicted key parts usually overlap with each other. In other words, there could be several confidence maps whose maximum values share the same location, introducing redundancy that wastes model parameters. Therefore, we additionally introduce a uniqueness loss to make sure that the predicted key parts do not share the same location.

The uniqueness loss dictates that the spatial maximum of the sum of confidences across all the maps on a foreground example should only be $1$. More specifically, for each local part on a foreground object, we first compute the channel-wise sum of confidence values for all maps. Then, we make the spatially maximum summed confidences as close to $1$ as possible. In this way, if two key parts with high confidence values overlap with each other on the same part, the sum of their confidence values on that part will easily exceed 1 and then the corresponding confidences will be penalized.
%The uniqueness loss will be penalize the confidence values on that part.
Suppose $c^i_s$ is the sum of confidence values across $K$ maps for the $i$-th object, and $(c^i_{s})_{m}$ is the spatial maximum summed confidence. Then, we have $\mathcal{L}_u$ as:
\begin{equation}
\mathcal{L}_u =\sum_{i}  \hat{y}^{i} \cdot l\big[ (c^i_{s})_{m}, 1 \big].
\end{equation}
Similar to Eq. \ref{eq:ld}, we also use the smooth L1 loss to define $l$ in $\mathcal{L}_u$. 

\vspace{-0.1cm}
\subsection{Condensing Two-stage Detection Head}
\label{sec:3-3}
\vspace{-0.1cm}
After predicting object key parts, we then condense the two-stage detection heads of the second stage.
% In general, the common two-stage detection head as described in Eq. \ref{eq:orig} will introduce excessive parameters to model all the information provided by $X$. 
As mentioned previously, we decompose the object appearance modeling into a key part modeling process which produces $Z_{K}$ and a global modeling process which produces $Z_{G}$. 

\textbf{Key Part Modeling}
We introduce key part modeling to extract semantics from detailed visual features of the discovered key parts. This allows us to obtain essential fine details to achieve high accuracy without introducing extra parameters to also model trivial features from non-key parts. 

In key part modeling, we first collect key part locations and then unify the features of these key parts. We denote $p_k$ as the spatial coordinates of the $k$-th discovered key part with the maximum confidence on the $k$-th map. We also denote $P$ as the collection of $p_k$ for all the $K$ key parts: $P=\{p_1, \ldots, p_k, \dots, p_K\}$. We use $\mathbf{x}_p$ to denote visual features from $X$ at $p$. Suppose $f_{OKPD}$ is the function that represents OKPD-Net, and we have:
\begin{equation}
Z_{K}=f_{K}(X, W_{K}) = g(\{\mathbf{x}_p | p \in P\}), ~~~ P = f_{OKPD}(X, W_{OKPD}),
\end{equation}
where $g$ is the function that unifies the collected $\mathbf{x}_p$ into the same tensor, and $W_{OKPD}$ are the OKPD-Net parameters. It is worth noting that $W_{OKPD}$ is equivalent to $W_{K}$ in this case. 
%In practice, we re-use the features of concentration step to refine $X$. Besides, 
In practice, we implement $g$ as a concatenation operation, during which we also include the confidence maps of key parts. One factor that should be of concern is how to arrange the collected $\mathbf{x}_p$ for unification. We find in our experiment that randomly distributing $\mathbf{x}_p$ during concatenation is inadequate, because the random key part features can confuse the detection and thus affect the final performance. To avoid this, we distribute $\mathbf{x}_p$ according to the orders of the corresponding key parts.

\textbf{Global Modeling}
Besides key parts, it has been found \cite{tanaka1993parts} that the biological visual system which also perceives objects as a whole. We find that a global modeling that extracts semantics from the overall appearance is necessary to help maintain the detection performance. %We achieve this by modeling the rough and holistic appearances and states of objects.
%To this end, we tend to introduce another coarse modeling process to model the general appearance about input objects. 

In global modeling, we model the rough overall appearances by sub-sampling the input features both spatial-wise and channel-wise. In particular, $Z_{G}$ is computed by:
\begin{equation}
Z_{G} = f_{G}(X, W_{G}) = h_c( h_{sp}(X), W_{G}),
\end{equation}
where $h_{sp}$ is the spatial-wise sub-sampling function, $h_c$ is channel-wise sub-sampling function, and $W_{G}$ is the parameters for $h_c$. In practice, the spatial sub-sampling $h_{sp}$ can be easily and appropriately implemented by adaptive average pooling. We then implement $h_c$ by adopting a 1 by 1 convolution to greatly reduces the input channel number using parameters $W_{G}$. Empirically, we keep a quarter of the original channels for channel-wise sub-sampling. 
%We also re-use the features of concentration step to refine $X$.

%\subsection{Object Key Part Discovery Network}
%\subsection{Effects Different Design Choices}

\vspace{-0.2cm}
\section{Experiments}
\label{sec:4}
\vspace{-0.2cm}
We performed comprehensive experiments to study the detailed effects of the proposed technique on condensing two-stage detection heads. Two widely used datasets, PASCAL VOC \cite{everingham2010pascal} and MS COCO \cite{lin2014microsoft}, were used for training and evaluation. We mainly used ResNet50 and ResNet101 as our backbone networks. Since we can hardly find another similar and effective study for condensing two-stage detection heads, we compared our method to the baseline detection heads. We used mmdet \cite{mmdetection} as our experimental platform because it provides implementation of many two-stage detectors. Performance was mainly measured by mean average precision (mAP). 
More training and testing details are provided in the appendices.
%More training and testing details will be provided later.

According to our method, the parameters for decomposing the object appearance modeling can determine the degree to which the model parameters will be condensed. For example, more key parts in key part modeling, or smaller sub-sampling rates in global modeling, can both result in more model parameters and \textit{vice versa}. Here, we mainly adjusted the number of key parts, denoted by ``K'', and sub-sampled spatial length, denoted by ``L'', to control the extent of reduction of model parameters. For example, ``K16,L5'' represents that we discovered and modeled 16 key parts and sub-sampled object appearance into the tensor of 5 by 5 spatial size. 
%, and the sub-sampled channel number, and 64 channels

\vspace{-0.1cm}
\subsection{Discovered Object Key Parts}
\begin{figure}
%\begin{floatrow}
%\ffigbox{%
\begin{minipage}{0.48\textwidth}
\includegraphics[width=7.0cm,height=4.3cm]{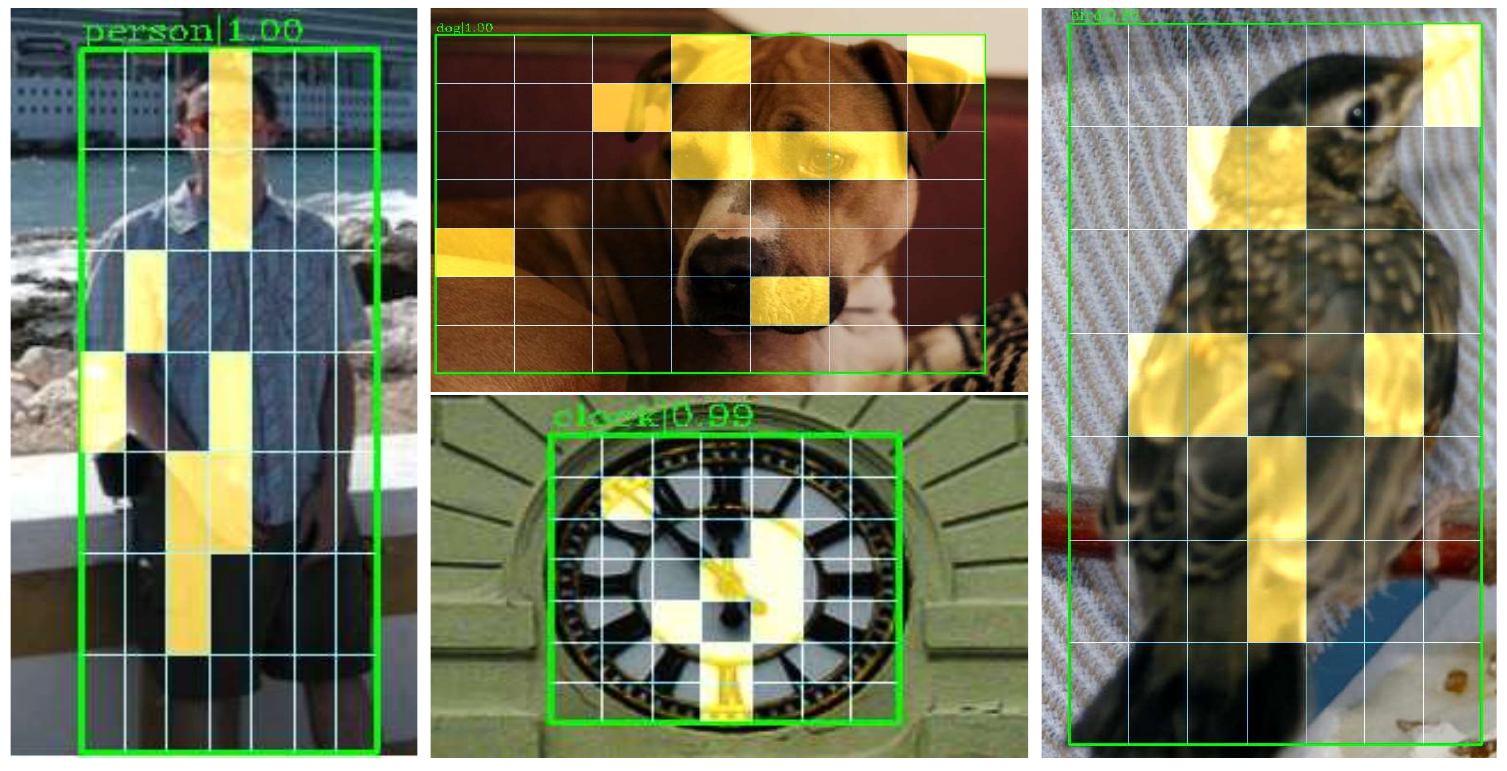}
  \caption{Discovered object key parts. 8 most confident discovered parts are presented as yellow boxes. Best view in color. }%
  \label{fig:1}
  \end{minipage}
  \hspace{0.5cm}
  \begin{minipage}{0.48\textwidth}
\includegraphics[width=6.8cm,height=4.7cm]{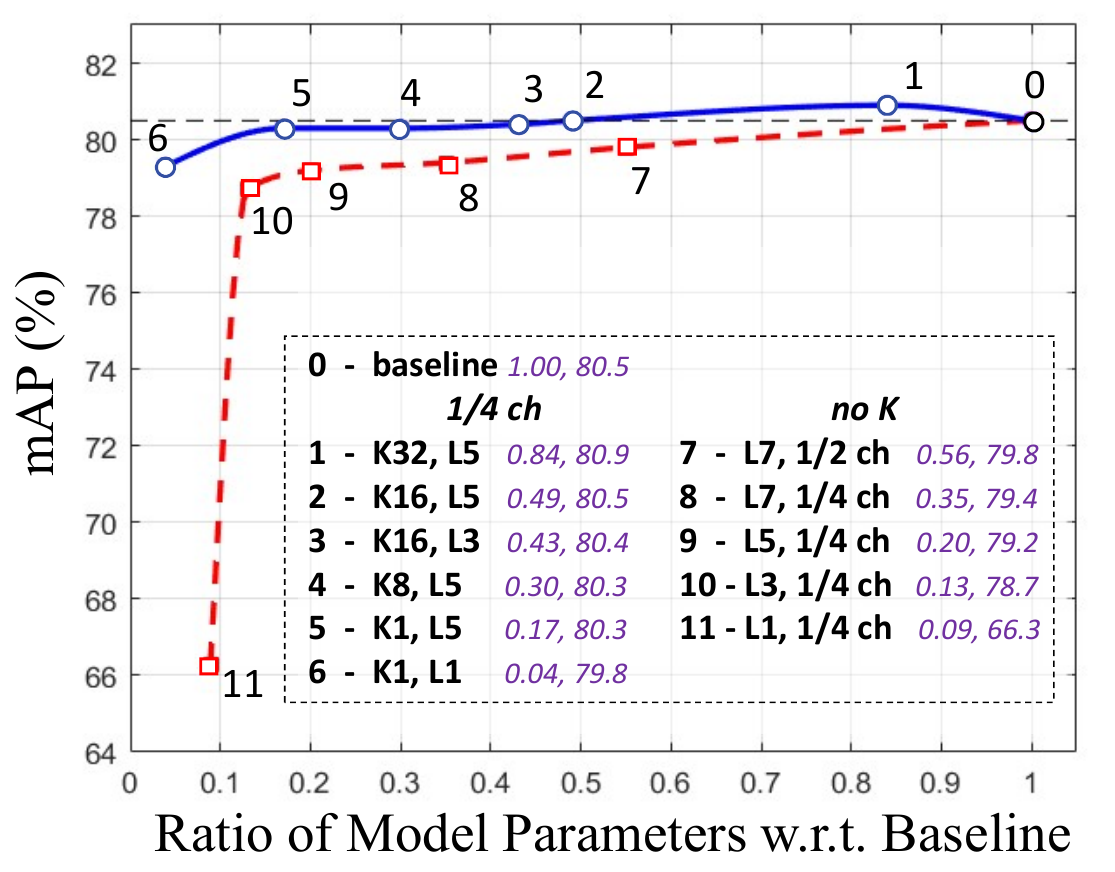}
  \caption{Effects of using key parts for condensing and reducing model parameters. "ch" means the original feature channel number.}%
  \label{fig:2}
  \end{minipage}
\vspace{-0.5cm}
\end{figure}

\vspace{-0.1cm}
We first present the automatically discovered object key parts for different objects. Figure \ref{fig:1} shows 8 out of 16 key parts with highest confidence scores for each object. According to the figure, it has been shown that the discovered key parts actually cover the more representative parts of objects. For example, in the left-most figure with a person inside, the discovered key parts include his head, body, and legs rather than background areas. Besides detection, the discovery of key parts can also help inspect the behaviors of neural networks and make them more explainable for analysis. 
More examples are provided in the appendices.
%More examples will be provided later.

%On PASCAL VOC datasets, we mainly study the effects of different components in our method and the effects of different design choices. On MS COCO dataset, the effects on different 

%\subsection{Discussion on effects of different design choices}
\vspace{-0.2cm}
\subsection{PASCAL VOC}
\vspace{-0.2cm}
Next, we performed experiments on the PASCAL VOC dataset. The ``07 + 12'' dataset was used for training and ``07'' test set was used for evaluation.

\begin{table}[b]
\begin{minipage}{0.45\textwidth}
  \caption{Comparison of different key part prediction methods. "TMR": truncated maximum regularization.}
  \label{tab:1}
  \centering
  \begin{tabular}{|l|c|} 
  \hline
  Key Part Prediction Method & mAP(\%) \\ \hline
  OKPD-Net + linear & 79.9 \\
  OKPD-Net + sigmoid & 79.8 \\
  OKPD-Net + softmax & 80.1 \\
   \hline
  OKPD-Net +  TMR (ours) & \textbf{80.5} \\ 
  \hline
  \end{tabular}
  \end{minipage}
  \hspace{0.5cm}
\begin{minipage}{0.45\textwidth}
  \caption{Ablation study of different components comparing to baseline methods. }%
  \label{tab:2}
  \centering
\resizebox{6.0cm}{!}{
  \begin{tabular}{|l|c|} 
  \hline
  Method & mAP(\%) \\ \hline
  baseline & 80.5 \\ \hline
  baseline pruning \cite{girshick2015fast,gong2014compressing} (halve model size) &  78.9 \\ \hline
  ours (halve model size: K16, L5) & \textbf{80.5} \\
  ours w/o concentration step & 80.0 \\ 
  ours w/o global modeling & 79.8 \\ 
  ours w/o $\mathcal{L}_d$ & 79.9 \\ 
  ours w/o $\mathcal{L}_u$ & 80.2 \\ 
  \hline
  \end{tabular}
}
  \end{minipage}
\end{table}

\textbf{Ablation Studies}
We first evaluated the effects of different components of our approach on condensing the detection head of the FPN \cite{lin2017feature} architecture which extracts visual features into a 7 by 7 tensor with 256 channels and
uses two fully connected layers to model appearance. ResNet50 was applied as the backbone network. 

Figure \ref{fig:2} presents the performance comparison for condensing the FPN detection head with or without object key parts. We can observe that condensing the two-stage detection by directly reducing channels or spatial sizes (red dotted line) can greatly abridge parameters at the cost of a significant degradation in accuracy. Conversely, introducing key part modeling (blue solid line) effectively reduced model parameters without obvious performance drop, even when using very few parameters for detection. In particular, with the ``K1, L1'' setting, our method waived over 96\% of the original model parameters with only a deterioration of 1.5 \% in performance.

Table \ref{tab:1} and Table \ref{tab:2} show ablation studies on different components introduced in our method, including the truncated maximum regularization, concentration step, global modeling, and training objectives $\mathcal{L}_d$ and $\mathcal{L}_u$. Baseline detection head and a baseline SVD-based network pruning technique \cite{girshick2015fast,gong2014compressing} are also evaluated for comparison. 
The results show that the compared pruning technique caused significant accuracy loss. Instead,
the proposed components are all beneficial to the performance when condensing two-stage detection. For example, $\mathcal{L}_d$ that ensures the quality of the discovered key parts contributed 0.6 mAP.

\textbf{Effects on Different Two-stage Detection Architectures}
%\label{exp}
We applied our method to condense different two-stage detection architectures, including Faster RCNN \cite{ren2015faster}, R-FCN \cite{dai2016r}, FPN \cite{lin2017feature}, and Cascade RCNN \cite{felzenszwalb2010cascade} . According to Figure \ref{fig:2}, our method fully preserved the high accuracy by with only 50\% of the original parameters. Therefore, we adjusted the parameters of our method to only waive around 50\% of parameters for different two-stage detection heads to validate whether our approach still retained their accuracy. ResNet101 was applied as the backbone network. Parameters for condensing different detectors can be found in the appendices.  
%Parameters for condensing different detectors will be provided later. 

Table \ref{tab:voc} shows the detailed results on the PASCAL VOC 07 test set. The number of parameters, model sizes, and model complexities (FLOPs in terms of GMAC)\footnote{FLOPs: floating point operations; GMAC: giga multiply–accumulate operations per second}, are illustrated. We can observe that our method consistently achieved comparable or even slightly better performance than baseline models using only half the original model parameters for detection. The model complexities are also promisingly reduced using our method.

\begin{table}[t]
\begin{center}
\caption{Performance and model scales of our method for condensing different detection heads in different two-stage detection architectures on VOC 07 test set.}
\label{tab:voc}
\resizebox{0.65\textwidth}{!}{
  \begin{tabular}{l | c | c | c | c   }
  \hline
  Detection Architectures & Params ($\times10^6$) & Size &  FLOPs (GMAC)  & mAP(\%) \\%& bike & bird & boat & bottle & bus & car & cat & chair & cow & table &  dog & horse & motor & person & plant & sheep & sofa & train & tv \\
  \hline
   R-FCN \cite{dai2016r} & 3.5 & 13 MB &  7.4 & 80.5\\
  Ours \& R-FCN& 1.5 & 5 MB & 4.8 & 80.6\\ %0.42
  \hline
   Faster RCNN \cite{ren2015faster} & 15.2 & 59 MB & 734.0 & 79.8 \\
  Ours \& Faster RCNN & 8.0 & 31 MB & 60.3 & 80.2 \\ %0.52
  \hline
   FPN \cite{lin2017feature}& 14.0 & 54 MB & 14.0 & 81.3 \\
  Ours \& FPN& 6.8 & 27 MB & 10.4 & 81.5 \\ %0.49
  \hline
  Cascade RCNN \cite{felzenszwalb2010cascade} & 41.8 & 160 MB & 41.8& 82.1 \\
  Ours \& Cascade RCNN & 20.3 & 78 MB & 34.8 & 82.0 \\ %0.49
  \hline
  \end{tabular}
}
\end{center}
\end{table}

\vspace{-0.1cm}
\subsection{MS COCO}
\vspace{-0.1cm}
MS COCO dataset \cite{lin2014microsoft} is larger and more complicated than PASCAL VOC. We therefore used it to further validate the effectiveness of our method for reducing model parameters for different two-stage detection architectures. We used the standard 1x schedule to train different detectors on MS COCO. Other settings are the same with the study \cite{lin2017feature}. ResNet101 was applied as backbone network.

Table \ref{tab:coco} shows the detailed results on MS COCO, which further confirmed that our method can adequately condense two-stage detectors. For example, the Faster RCNN detector performed detection using its entire last convolutional stage with 15.8 million parameters and a computational complexity of 733.8 GMac. By replacing its detection head, our method achieved slightly better accuracy while waiving over a half of the parameters and over 91\% complexity. Furthermore, our approach consistently delivered comparable detection performance to the state-of-the-art Cascade RCNN detector while waiving around a half of its mode parameters.

\begin{table}[!t]
\caption{Performance and model scales of our method for condensing different detection heads in different two-stage detection architectures on MS COCO val set.}
\label{tab:coco}
\begin{center}
\resizebox{\textwidth}{!}{
  \begin{tabular}{|l | c | c | c| c| c | c | c | c |}
  \hline
  Detection Architectures & Params ($\times10^6$)  & Size & FLOPs (GMAC) & AP & mAP@0.5 & mAP(small) & mAP(medium) & mAP(large) \\
  \hline  
  R-FCN \cite{dai2016r}  & 6.6 & 25 MB &  25.4 & 27.6  & 48.9 & 8.9 & 30.5 & 42.0\\
   Ours \& R-FCN & 3.0 & 12 MB & 7.8 & 28.4 & 46.0 & 7.3 & 30.8 & 47.9 \\%0.45 
  \hline    
  Faster RCNN \cite{ren2015faster} & 15.8 & 61 MB & 733.8 &37.4  & 58.7 & 18.8 &41.0 &51.1\\
   Ours \&  Faster RCNN &8.3 & 32 MB & 60.6 & 37.5 & 58.7 & 21.8& 41.9 & 48.6 \\ %0.46
  \hline
  FPN \cite{lin2017feature} & 14.3 & 55 MB &  14.3 & 38.5  & 60.3& 22.3 & 43.0 & 49.8\\
   Ours \& FPN & 7.1 & 28 MB &10.6 & 38.5 & 60.4 & 22.4& 42.8 &50.3 \\ % 0.50
  \hline
  Cascade RCNN \cite{felzenszwalb2010cascade} & 41.9 & 161 MB &  41.9 &42.0  & 60.3 & 23.2 & 45.9 & 56.3\\
  Ours \& Cascade RCNN & 20.5& 79 MB & 35.0 & 42.1& 61.1 &23.1& 46.0 & 57.0 \\ %0.49
  \hline

  \hline
  \end{tabular}
}
\end{center}
\vspace{-0.2cm}
\end{table}

\vspace{-0.2cm}
\section{Conclusion and Future Work}
\vspace{-0.2cm}
We introduced a novel approach to effectively condense and reduce the model parameters of two-stage detection heads using automatic object key part discovery. To our best knowledge, the proposed method is the first that can waive more than 50\% of the model parameters of original two-stage detection heads without sacrificing performance, illustrating that complicated detection models can be greatly simplified with the combination of global modeling and key part modeling. 

It is worth mentioning that our technology is orthogonal to simplifying or compressing backbone classification networks like ResNet\cite{he2016deep}, suggesting that the cooperation of both directions could further reduce parameters of the two-stage detectors. In the future, we plan to extend the idea of decomposing appearance modeling into a key part modeling process and a global modeling process to also recast and condense normal convolution operations and other computer vision tasks.

\section*{Broader Impact}
Object detection is the foundation of various computer vision tasks and applications like video analysis and autonomous driving. Our research can significantly save model parameters required by high-quality two-stage detectors without sacrificing much performance. Besides, the discovered key parts are beneficial to improve the interpretability of neural networks, facilitating researchers to better diagnose networks and analyze how they model visual patterns. 

In general, our research could benefit researchers and engineers from diversified areas. By making models more compact and more efficient, our research tends to promisingly lower the expenses of computation for accurate detection, providing a powerful tool to ease the difficulty of applying cutting-edge two-stage detectors to various tasks and platforms. Meanwhile, with a more computationally friendly high-quality detector, our research could also save great labor and energy costs for the tasks like traffic monitoring, vehicle navigation, social media analysis, \textit{etc.}, creating more opportunities to apply deep learning in the real-world and to improve the productivity of human society.

However, by providing a more efficient accurate detection model, our research could be abused to put peoples' privacy at risk. For example, the combination of an efficient and powerful pedestrian detector and a face recognizer can be applied in security cameras to identify some specific people and monitor their daily lives without permission. To avoid such negative impacts, appropriate regulations should be taken into account in the research community and industrial companies. 

\section*{Appendix A: Settings for Training and Testing}
We describe in this section more details about the training and testing settings of our proposed approach used in our ``Experiment'' section for condensing two-stage detection with automatic object key part discovery. 

\subsection{PASCAL VOC}
PASCAL VOC \cite{everingham2010pascal} is a popular detection dataset that has 20 object categories. We use the union set of ``VOC 07'' \textit{trainval} and ``VOC 12'' \textit{trainval} for training. This union set has around 20,000 images. We then use the test set of ``VOC 07'' for testing. This test set has around 5,000 images.

When training two-stage detection models with our approach, we use 12 training epochs to optimize network parameters. We use a batch size of 8 with a learning rate of 0.01. We decay the learning rate at 10-th epoch by multiplying it with 0.1. We resize training images such that their shorter sizes are 600 and longer sizes do not exceed 1000. 
When testing the trained two-stage detectors with our approach, we also resize images in a similar way we perform in the training phase. No augmentation techniques are used in training and testing.

\subsection{MS COCO}
MS COCO\cite{lin2014microsoft} is a widely used object detection dataset that has 80 object categories. It has around 83,000 images for training. We uses its \textit{val} set, which has 5,000 images, for testing. This dataset is much more challenging than PASCAL VOC because it contains many small and obscure objects. 

As mentioned in the paper, we use a similar training setting to FPN \cite{lin2017feature}. Both the training and testing images are resized so that their shorter sizes are 800 and longer sizes do not exceed 1333. No augmentation techniques are used in training and testing. To ensure the original detection performance on MS COCO dataset, we tend to first use original training setting to pre-train the backbone network and then use our method to condense two-stage detection.

\subsection{Settings for Condensing Different Detection Architectures}
Here, we describe the details of how we use our method to replace the two-stage detection heads as applied in R-FCN, Faster RCNN, FPN, and Cascade RCNN for the experiments conducted in Table 3 and Table 4 in the paper. 

\textbf{R-FCN} The original R-FCN uses the entire backbone network for feature extraction, but its last convolutional stage does not perform down-sampling. Therefore, its output feature maps have a spatial size 16 times smaller than the input images. In its detection head, it applies a 1 by 1 convolution to reduce the channel number from 2048 to 1024, followed by another convolution to predict classes and spatial offsets. For more details, please refer to \cite{dai2016r}.

To condense the detection head of R-FCN using our method by waiving around 50\% of original model parameters, we first keep the settings of the backbone network. Then, rather than reducing the channel number as in original R-FCN, we reduce the channel number to 256, and apply ``K16, L3'' setting for condensing. We do not introduce the fully-connected layer in this case. 

\textbf{Faster RCNN} The original Faster RCNN uses the first 4 convolutional stage of backbone network for feature extraction, and uses its entire last convolutional stage as detection head. Its output feature maps also have a spatial size 16 times smaller than the input images. For more details, please refer to \cite{ren2015faster}.

Similar to R-FCN, we keep the original backbone network setting to condense the detection head of Faster RCNN. Therefore, the channel number of extracted features for detection is 1024. Instead of using the entire last convolutional stage, we apply the proposed detection head with a setting of ``K4, L3'' and apply 4 convolutions in the concentration step of OKPD-Net rather than 2 convolutions as normally applied. This result in 50\% less model parameters while still maintaining original performance.

\textbf{FPN} The original FPN uses the entire backbone network and also an additional convolutional stage to perform detection at scales from 4 times smaller than input images to 64 times smaller. All the channel numbers of extracted features for detection from different levels are reduced to 256. For more details, please refer to \cite{lin2017feature}.

To condense FPN, we apply the ``K16, L5'' setting to waive around a half of original model parameters. Other settings are as described in the paper. 

\begin{figure}[t]
\includegraphics[width=\textwidth]{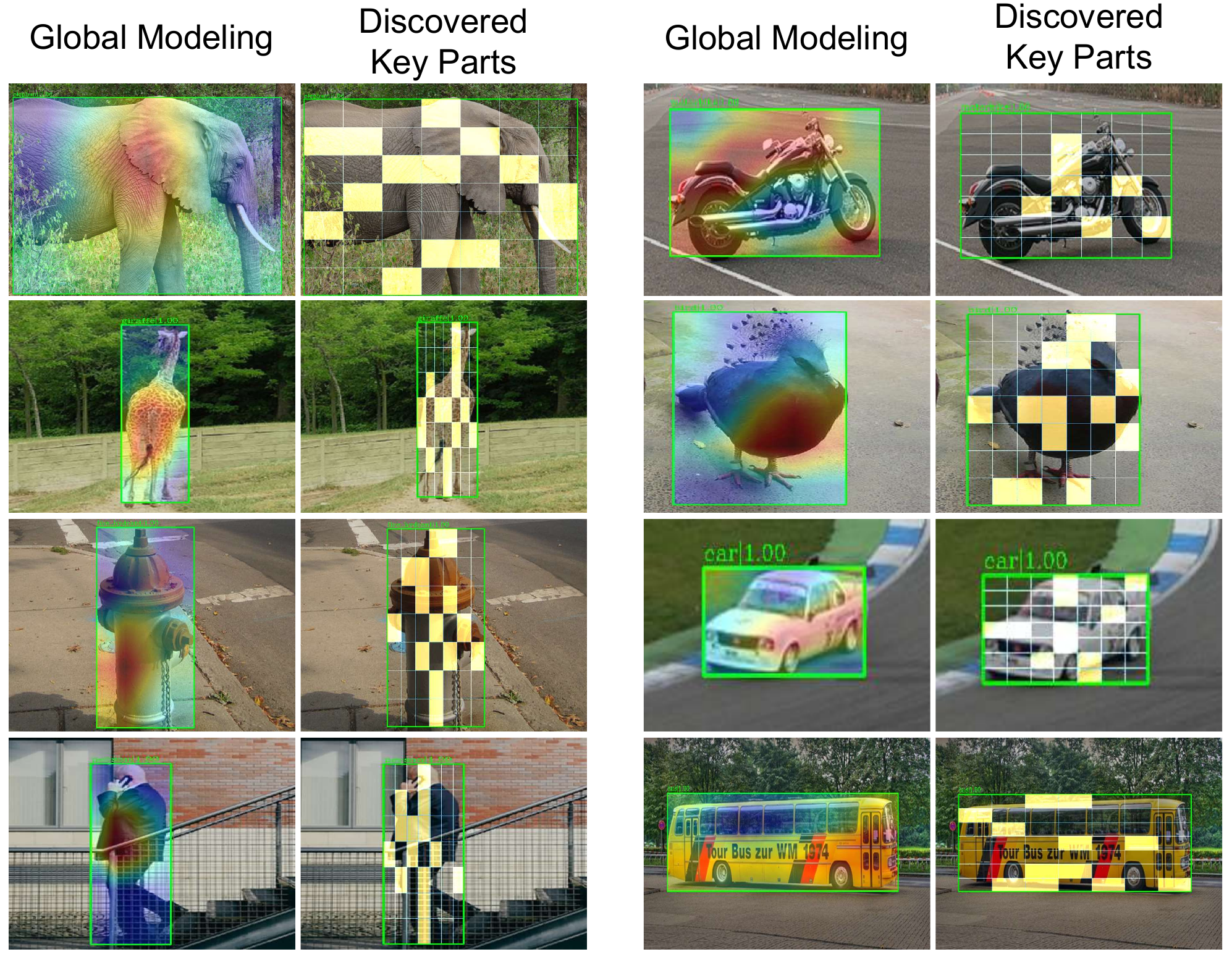}
  \caption{More qualitative results. We present discovered key parts and feature activation maps of the outputs of global modeling process.}%
  \label{fig:qua-2}
  \vspace{-0.5cm}
\end{figure}

\textbf{Cascade RCNN} The original Cascade RCNN follows a similar backbone setting as FPN but with 2 additional detection heads to iteratively refine detection results. For more details, please refer to \cite{felzenszwalb2010cascade}.

Condensing Cascade RCNN is similar to condensing FPN. The only difference is that we apply 3 and 4 convolutions for the concentration step in OKPD-Nets of the additional two detection heads, respectively.

\section*{Appendix B: More Qualitative Results}
We present more qualitative results of both global modeling and key parts discovery in Figure \ref{fig:qua-2}. 

In the figure, we first draw feature activation map for global modeling using heat maps. In particular, for the output features of global modeling process, we first squash the values with ReLU, and then compute the sum of all feature activations. Afterwards, we normalize the values to lie in the range of [0,1], and the normalized values are transformed into a heat map where red areas denote high activations and blue areas denote low activations. According to the presented results, we can find that global modeling preserves the general spatial arrangement of objects in the proposal by producing high values on representative areas of objects, \textit{e.g.} the center areas of elephants, birds, cars, and people. It indeed focuses more on the general representation and most discriminative areas of objects. 

Besides global modeling, we also present dicovered key parts in the figure. For the key parts, we present the ones with confidences which are larger than 0.1. According to the illustrated key parts, we can easily confirm that the key parts better cover the representative detailed areas of objects. For example, to detect a giraffe, the key parts focus on its neck and legs, while global modeling simply models its body.
Therefore, the features from key parts can provide more details that facilitates the accurate detection of objects. Besides, the presented key parts can better align the bounding boxes and sample more on areas where global modeling has low activations, meaning that key part scan effectively complement global modeling.  

Overall, the presented results demonstrate that key part modeling provides crucial and complementary information in addition to global modeling, thus can allow the significant reduction of the model parameters without sacrificing much performance. 

{\small
\bibliographystyle{unsrt}
\bibliography{egbib_nips}
}
\end{document}